\def\year{2021}\relax
\documentclass[letterpaper]{article} 
\usepackage{aaai21}  
\usepackage{times}  
\usepackage{helvet} 
\usepackage{courier}  
\usepackage[hyphens]{url}  
\usepackage{graphicx} 
\usepackage{natbib}  
\usepackage{caption} 
\nocopyright
\title{An Investigation of the Weight Space \\ to Monitor the Training Progress of Neural Networks}
\author{
    Konstantin Sch\"{u}rholt,~
    Damian Borth,
    \\
}
\affiliations{
    University of St.Gallen, Switzerland\\

    konstantin.schuerholt@unisg.ch

}

\begin{document}

\maketitle
%
%
%
\begin{abstract}
Safe use of Deep Neural Networks (DNNs) requires careful testing. However, deployed models are often trained further to improve in performance. As rigorous testing and evaluation is expensive, triggers are in need to determine the degree of change of a model.
In this paper we investigate the weight space of DNN models for structure that can be exploited to that end. Our results show that DNN models evolve on unique, smooth trajectories in weight space which can be used to track DNN training progress.
We hypothesize that curvature and smoothness of the trajectories as well as step length along it may contain information on the state of training as well as potential domain shifts.
We show that the model trajectories can be separated and the order of checkpoints on the trajectories recovered, which may serve as a first step towards DNN model versioning.
\end{abstract}

%
%
%
\section{Introduction}
\label{introduction}

\begin{figure}[t]
\begin{center}
\centerline{\includegraphics[width=0.95\columnwidth]{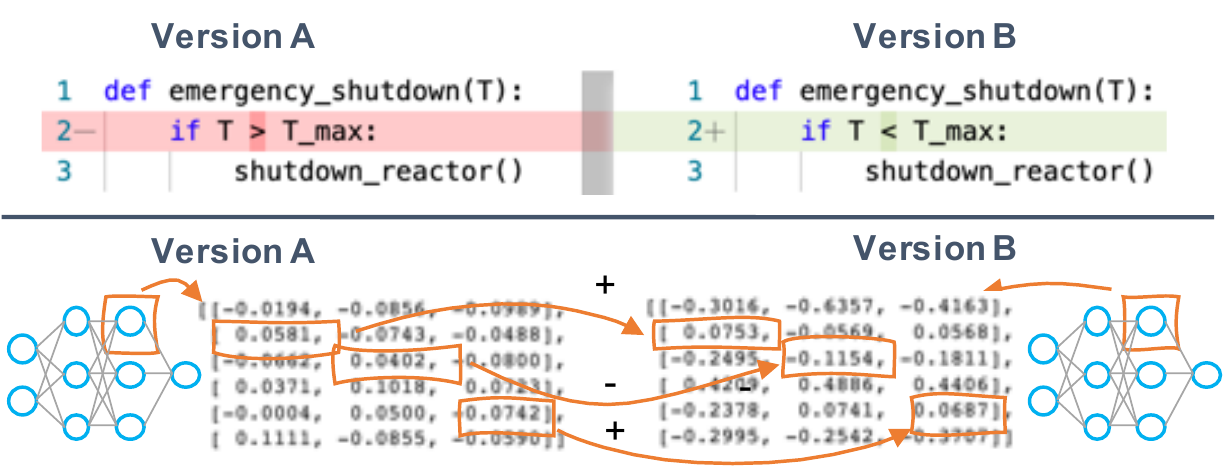}}
\caption{Illustration of the differences between traditional code and DNN models with respect to changes. \textbf{Top:} code changes immediately relate to behavior changes. \textbf{Bottom:} weight changes are obscure by complexity.}
\label{fig:version_control}
\end{center}
\end{figure}

In recent years, Deep Neural Networks (DNNs) have evolved from laboratory environments to the state-of-the-art for many real-world problems.
With that, the safety, neutrality and reliability requirements have risen.
To that end, DNN safety is usually determined by rigorous testing, i.e. for model biases or by applying corner cases.
These tests can be performed for regular major release cycles, but pose challenges for continual learning systems. Such systems, which keep learning as new data is collected, require measures to trigger testing and accreditation.
The usual training performance metrics, i.e. loss, accuracy, f1-scores, don't necessarily reveal changes that are relevant for testing, i.e. a learned class-bias.
Conventional software versioning methods may be used to track models, but as all of a model's parameters may change in a single epoch, regular diffs between versions aren't helpful without the context of changes (see Figure \ref{fig:version_control}).
%
They offer generally very little intuition into how far or in which direction a model has changed, whether changes are in-line with past development or abnormal and require further attention.
We therefore investigate the weight space of DNNs for properties that may be exploited for DNN monitoring.
In particular, we investigate the trajectories of DNN models through weight space during training. Our key findings are:
\begin{figure*}[t!]
\begin{center}
\centerline{\includegraphics[width=1.85\columnwidth]{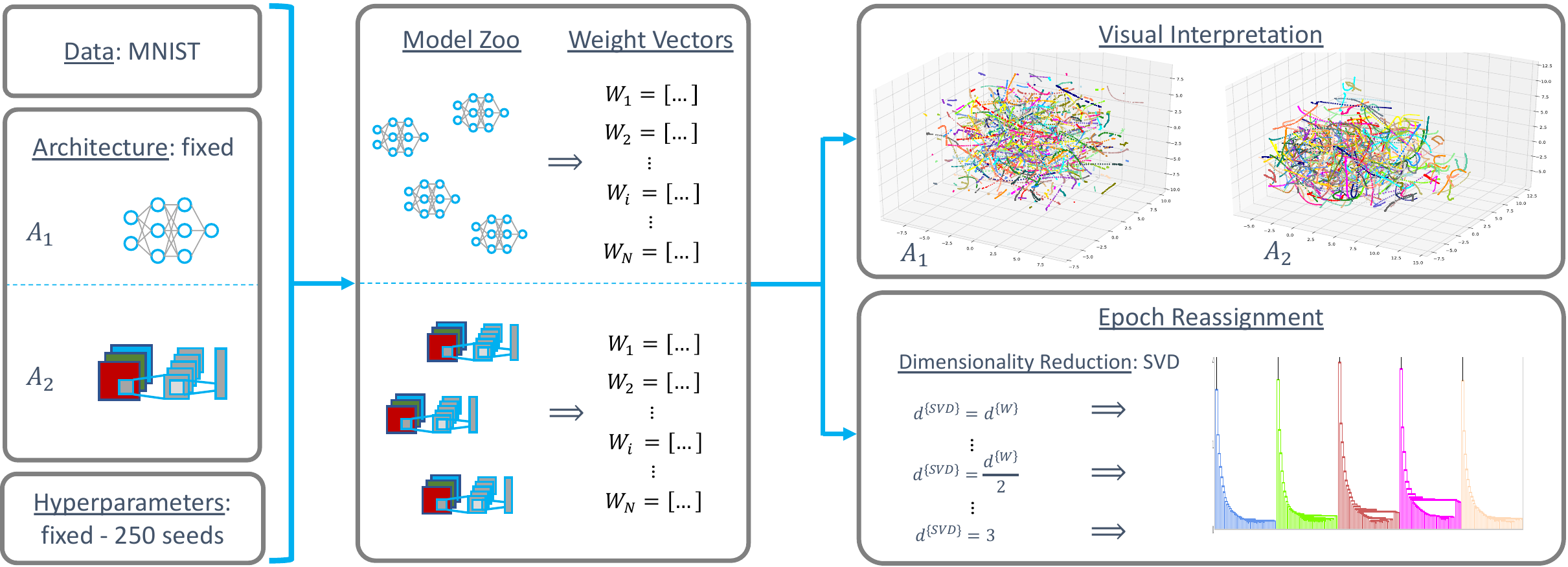}}
\caption{Overview of our approach to exploit weight space structure for DNN model change tracking}
\label{fig:approach_scheme}
\end{center}
\end{figure*}
%
\begin{itemize}
    \item Due to the high dimensionality of model weight spaces, even those of popular architectures are sampled only very sparsely. Moreover, the changes in parameter we encounter is rather local, i.e. trajectories do not cross the entire weight space. Therefore, model trajectories are quasi-unique, they are very unlikely to cross each other,  and so model trajectory clusters can be fully recovered using single-linkage agglomerative clustering.
    \item Model trajectories are smooth, which is unsurprising considering usual training methods applying momentum. Lack of smoothness or changes in curvature of trajectories may serve as indicators for a domain shift in the data, which may serve as indicators to investigate further.
    \item In our experiments, the step length along the trajectories mostly relates to the state in training. Initial updates in one epoch are larger, the updates get smaller as the model converges. The distance between updates is sufficient to recover order of model checkpoints on a trajectory to over 70\%. While the relation between loss and update is intuitive at least in the beginning of training, the progress along a trajectory averaged over an epoch may be utilized to track training progress even if the loss stalls.
    \item Both sparsity and order of model checkpoints on trajectory are surprisingly stable when the weight space is compressed, e.g. using singular value decomposition.
\end{itemize}

The remainder of this paper is structured as follows. After related work we present our approach and close with qualitative visualizations of model similarities and quantitative evaluation of the topological structure of the weight space.

\section{Related Work}
\label{sota}
%
%
In software engineering ``Git''~\cite{git_git_2020} has become the de-facto standard for version control and tracking of deployed software code into production~\cite{rhodecode_version_2017, synopsis_compare_2020}.
However, there are shortcomings by treating DNN models as software code as discussed in~\cite{miao_towards_2017}. The authors propose the concept of a ``modelhub''~\cite{modelhub_modelhub_2020} to track the process of design, training and evaluation of DNN by storing data, models and experiments. Similarly, tools like DVC~\cite{dvc_open-source_2020} and ``Weights \& Biases''~\cite{wandb_weights_2020} do follow the same direction with the aim to increase transparency and reproducibility of the model development pipeline. Unfortunately, since these tool strongly rely on the availability of manually generated human annotation (commit messages) to illustrate model changes, they are limited with respect to an automated and consistent deployment monitoring of DNN models.
%

%
%
With respect to DNN model analysis, we see two groups of research directions: 1) methods to compare two models, measuring their similarity; 2) absolute methods which map one model to an understandable space. Both methods can be either based on the model's parameters (weight space) or the activations given certain input (activation space). For the latter, the choice and availability of input data are critical.

\textbf{Relative Metrics} compare two models and compute a degree of similarity. As DNNs are graphs, graph similarity metrics can be applied, e.g.~\cite{soundarajan_guide_2014,mheich_siminet:_2018}.
In~\cite{laakso_content_2000}, the authors compare the activations of Neural Networks as a measure of "sameness". In~\cite{li_convergent_2015}, the authors compute correlations between the activations of different nets.~\cite{wang_towards_2018} try to match subspaces of the activation spaces of different networks, which~\cite{johnson_subspace_2019} show to be unreliable. In~\cite{raghu_svcca:_2017, morcos_insights_2018, kornblith_similarity_2019}, correlation metrics with certain invariances are applied to the activation spaces of DNNs to investigate the learning behavior or compare nets.

\textbf{Absolute Metrics} map single models to an absolute representation.~\cite{jia_geometric_2019} approximate a DNN's activations with a convex hull.~\cite{jiang_predicting_2019} also uses the activations to approximate the margin distribution and learn a regressor to predict the generalization gap. In~\cite{martin_traditional_2019} the authors relate the empirical spectral density of the weight matrices to model properties, i.e. test accuracy, indicating that the weight space alone contains a lot of information about the model. This is underlined by two recent publications, which generate large datasets of sample models and learn larger models to predict the sample model's properties from their weight space. In~\cite{eilertsen_classifying_2020}, the authors predict training hyperparameters from the weight space.~\cite{unterthiner_predicting_2020} learn a regressor to predict the test accuracy from the weight space.
%
%
%
%
\section{Approach}
\label{approach}
While relative metrics such as (SVCCA~\cite{raghu_svcca:_2017,morcos_insights_2018} and CKA~\cite{kornblith_similarity_2019}) have shown to be promising in providing similarity between DNN models, they depend on expressive data for model comparison. A comparison of two models by their mapping does not scale or generalize to a comprehensive representation of the model space. More specifically, in a transfer-learning or fine-tuning setup, individual models quickly become self-dissimilar with SVCCA or CKA, while different models evaluated on the same data could not always be told apart.
\begin{table*}[t]
\caption{Summary of Model Zoo Setup for MLP \textbf{(left)} and CNN \textbf{(right)} architectures}
\label{tab:zoo}
\begin{center}
\begin{tabular}{l|l}
\multicolumn{2}{c}{\textbf{Architecture Configurations}}       \\
($A_1$): MLP & ($A_2$): CNN \\
fc $\to$ act $\to$ fc $\to$ act $\to$ fc  &  $2 \times$ [conv $\to$ max-pool $\to$ act] $\to$ fc \footnote{fc: fully connected linear}\\
Input dim.: 196 (flattened greyscales) & Input dim.: 14x14 (greyscales) \\
Hidden layer: 10 units & Conv: 16 channels, stride 2, kernel size 3 \\
Activation (act): sigmoid & Activation (act): Leaky ReLU \\
Weight init: Kaiming Normal & Weight init: Kaiming Normal \\
\multicolumn{2}{c}{\textbf{$\lambda$ - Learning Setup}}       \\
Learning rate: 0.03 & Learning rate: 0.0003 \\
Momentum: 0.9 & Momentum: 0.9 \\
Weight decay: 0.0003 & Weight decay: 0.1 \\
Batchsize: 4 & Batchsize: 12 \\
\multicolumn{2}{c}{\textit{both architectures were trained with SDG on MSE}}       \\

\end{tabular}
\end{center}
\end{table*}
In contrast, the work from \cite{martin_traditional_2019,eilertsen_classifying_2020,unterthiner_predicting_2020} shows that the weight space already contains comprehensive information about DNN models. For this work, we aim to abstract from the weight space to track model evolution during training.
To that end, we aim to identify individual models trajectories. This allows us to tell different model apart and to link individual training epochs of the same model without further prior knowledge.
%
%
%
The approach is outlined in Figure~\ref{fig:approach_scheme}: drawn from two fixed architectures, we train a zoo of individual models forming a weight space of individual models, which is investigated visually and using automatic clustering to identify individual model training trajectories. In particular, the epoch-to-model assignment demonstrates its potential for tracking changes of DNN models.
\subsection{Model Zoo Generation}
We begin by training a ``model zoo'' of DNNs.
Similarly to~\cite{unterthiner_predicting_2020}, we denote $ S_{N}= \{ (\mathcal{X},\mathcal{Y})_{i} \}^{N}_{i=1}$ as the\textit{training set}, $\lambda$ as the set of hyper-parameter during training (e.g. loss function, optimizer, learning rate, weight initialization, batch-size, epochs), whereas we differentiate between specific \textit{neural architectures} drawn from a set of potential neural architectures denoted by $ A \in \mathcal{A}$. Further, during training the learning procedure $l(\cdot) \mapsto W $ which generated a DNN model $W$ depends on $S_{N}, A, \lambda$ and a fixed \textit{seed} specified apriori. Again, similar to~\cite{unterthiner_predicting_2020} $W$ denotes a flattened vector of model weights.

More specific, we train DNN models $W_i$ on $S_N$, the MNIST classification dataset\footnote{to keep models compact we have reduced image resolution to $14\times14$px} ~\cite{lecun_gradient-based_1998}, with two fixed architectures MLP and CNN and their corresponding $\lambda$, ref. Tab.~\ref{tab:zoo}. In total we fix $m=250$ seeds leading to two models zoos $\mathcal{W}_{A_1}$, and $\mathcal{W}_{A_2}$ with each $\mathcal{W} =  \{ W_1,...,W_{m} \}$ models. Please note the slight difference  to~\cite{unterthiner_predicting_2020}. While~\cite{unterthiner_predicting_2020} covers a large space of hyper-parameters $\lambda$ with one seed per $\lambda$, in this work, we have one fixed hyper-parameter $\lambda$ and train with a range of different seeds. Each model $W_i$ represents a sequence of weights $W_i=\{W^1_i, ...,W^k_i \}$ acquired for each epoch during training.
\subsection{Investigation of Weight Space Structure}
\label{approach:structure}
We first visually inspect the weight space with UMAP~\cite{mcinnes_umap_2018} and Singular Value Decomposition (SVD), which we use for dimensionality reduction of the weight space.
To automatically identify learning trajectories  $W_i=\{W^1_i, ...,W^k_i \}$, we apply agglomerative clustering on the full weight space and SVD reductions of it. We use single-linkage distance, as it accounts best for the 'string of pearls' trajectories, where one epoch is closest to the end of the string of previous epochs.
To quantify the results, we use a cut-off distance so that 250 cluster are identified. In these, we compute purity as $\frac{\textit{samples from the dominant net}}{\textit{samples in the cluster}}$. We further attempt to reconstruct the epoch-order. To that end, we order the samples of the dominant DNN model in each cluster by decreasing distance. We evaluate the number of total ordered epochs starting from the first epochs as well as the overall percentage of ordered epoch-to-epoch sequences.
%

\begin{figure}[t]
\begin{center}
\centerline{\includegraphics[width=1.1\columnwidth]{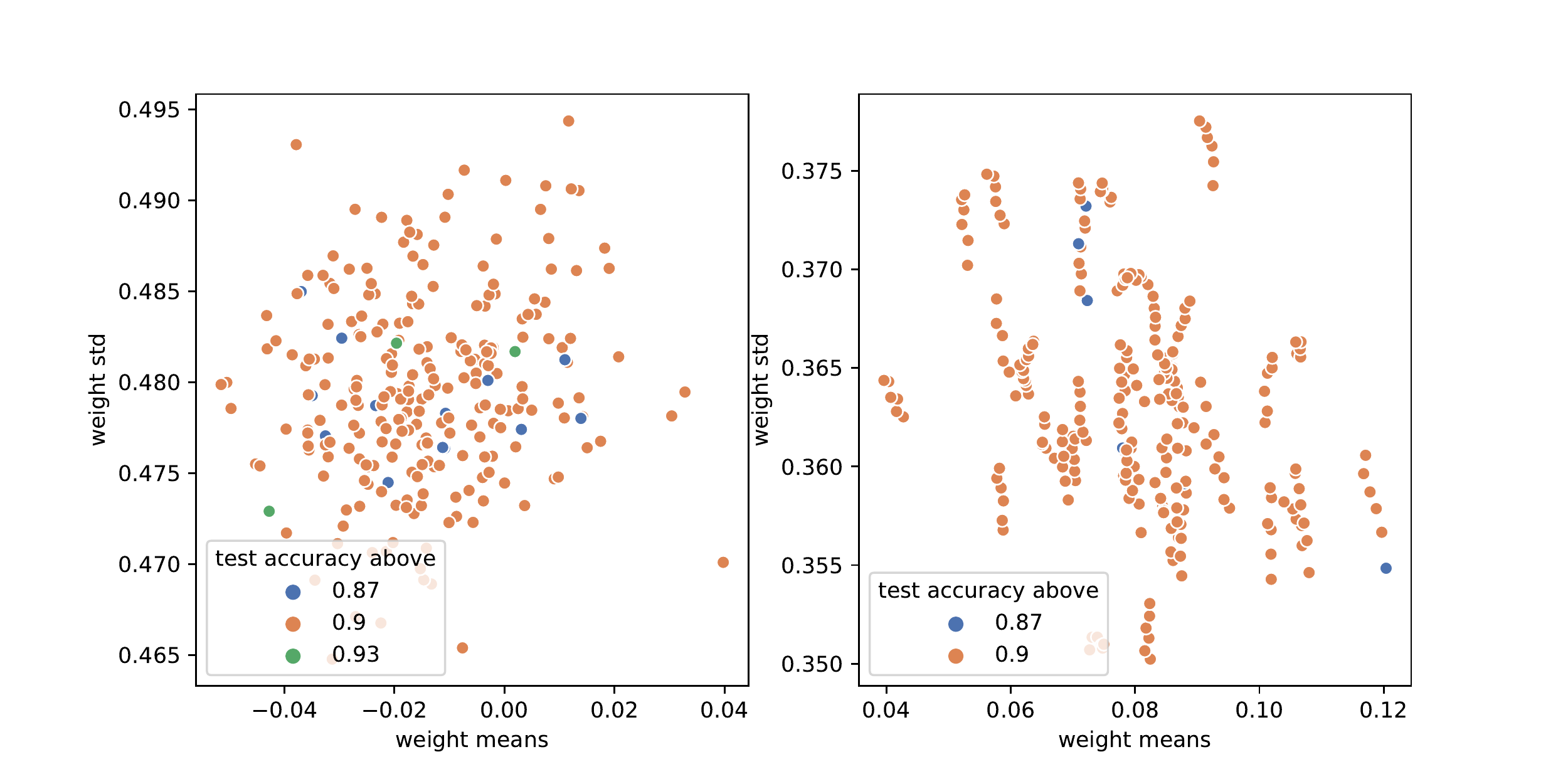}}
\caption{\textbf{Left}: Weight distribution of CNN models (epoch 50). \textbf{Right}: weight distribution of the MLP models (epoch 50). Each point represents the mean and standard deviation of all weights defining one DNN model. Colors indicate test set accuracy.}
\label{fig:weight_distro}
\end{center}
\end{figure}
%
%
\section{Results}
\label{results}
%
\begin{figure*}[t]
    \centering
    \centerline{\includegraphics[width=2\columnwidth]{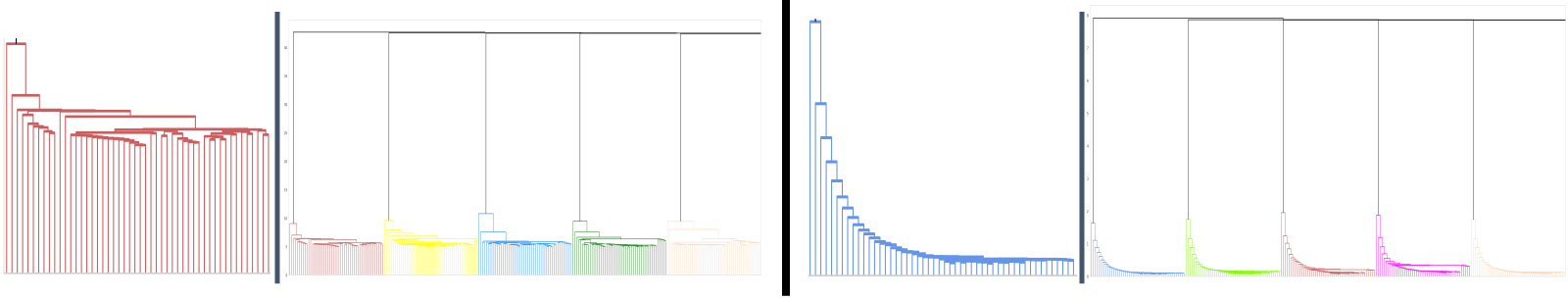}}
\caption{Result of the agglomerative clustering visualized as a dendrogram. \textbf{Left}: MLP, \textbf{right}: CNN. Epochs of the same DNN model, which are correctly clustered are colored in the same color. Only 5 out of 250 models are shown, where one cluster is additionally plotted in more detail.}
\label{fig:dendrograms}
\end{figure*}
%
We first evaluate the performance and weight distribution of the model zoos for each architecture independently. In Figure \ref{fig:weight_distro} every point represents the mean and standard deviation of all weights comprising the DNN model respectively, whereas colors indicate testset accuracy of the model. Both model zoos form distributions with very similar mean and standard deviation.
For MLP models, the range of weight means and standard deviation is without recognizable pattern while for CNN models the mean and standard deviation shows structure. Please note, that the ranges of the means are constrained by the used activation function as seen in~\ref{tab:zoo}. For both architectures, classification performance is as expected competitive given the simple classification task but not limited to a specific range of weight means and deviation deviations.
The difference in initialization seed appears not to meaningfully affect the distribution of the weights. By design, the individual models cannot be reliably told apart in this metric. This qualifies the zoos for our further purposes.
\subsection{Weight Space Visualization}
\begin{figure}[ht]
    \centering
    \centerline{\includegraphics[width=\columnwidth]{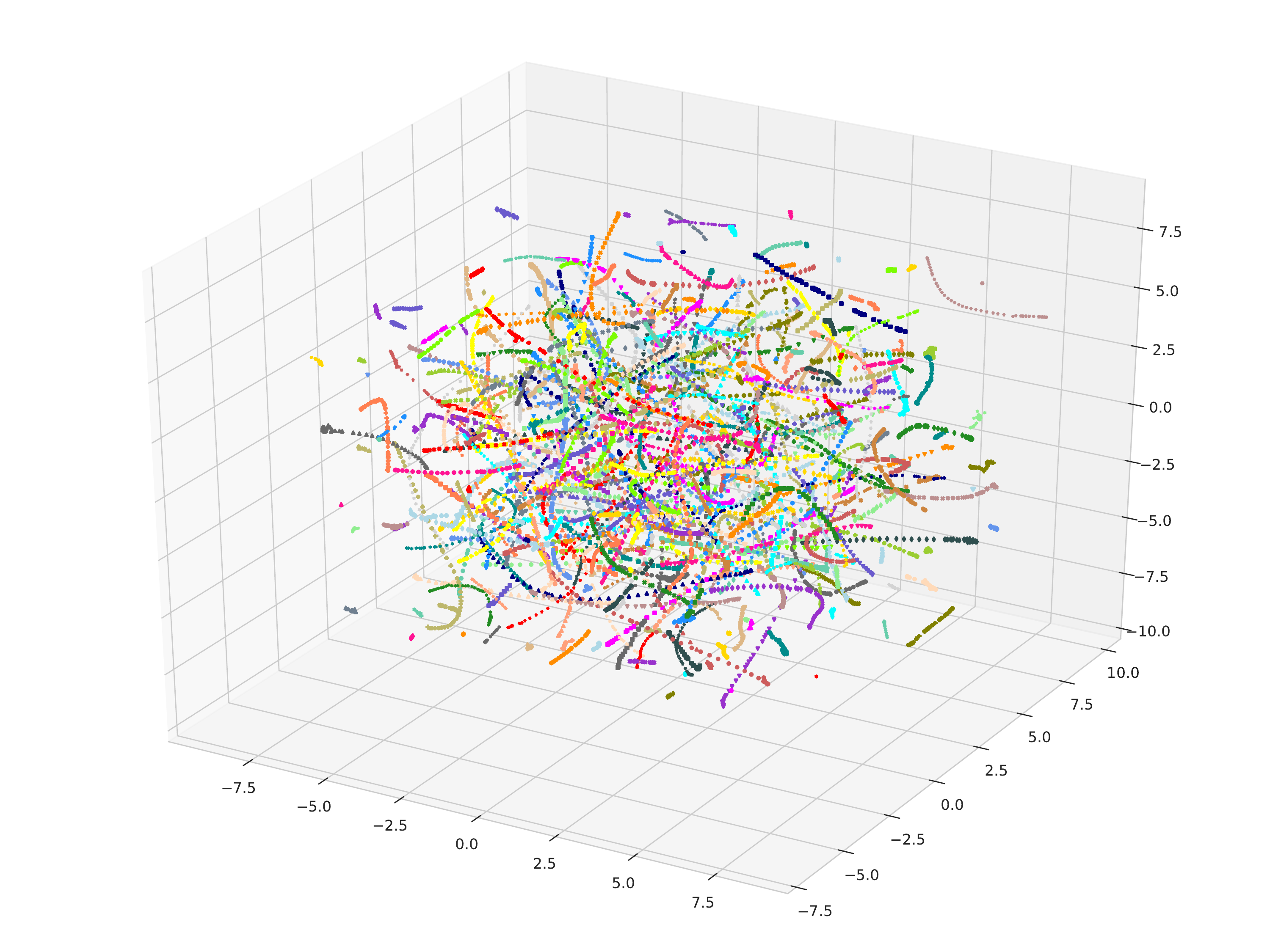}}
    \caption{UMAP of MLP weight space. Points of the same color represent the same network at different epochs. }
    \label{fig:weight_space_MLP_umap}
\end{figure}

\begin{figure}[ht]
    \centering
        \centerline{\includegraphics[width=\columnwidth]{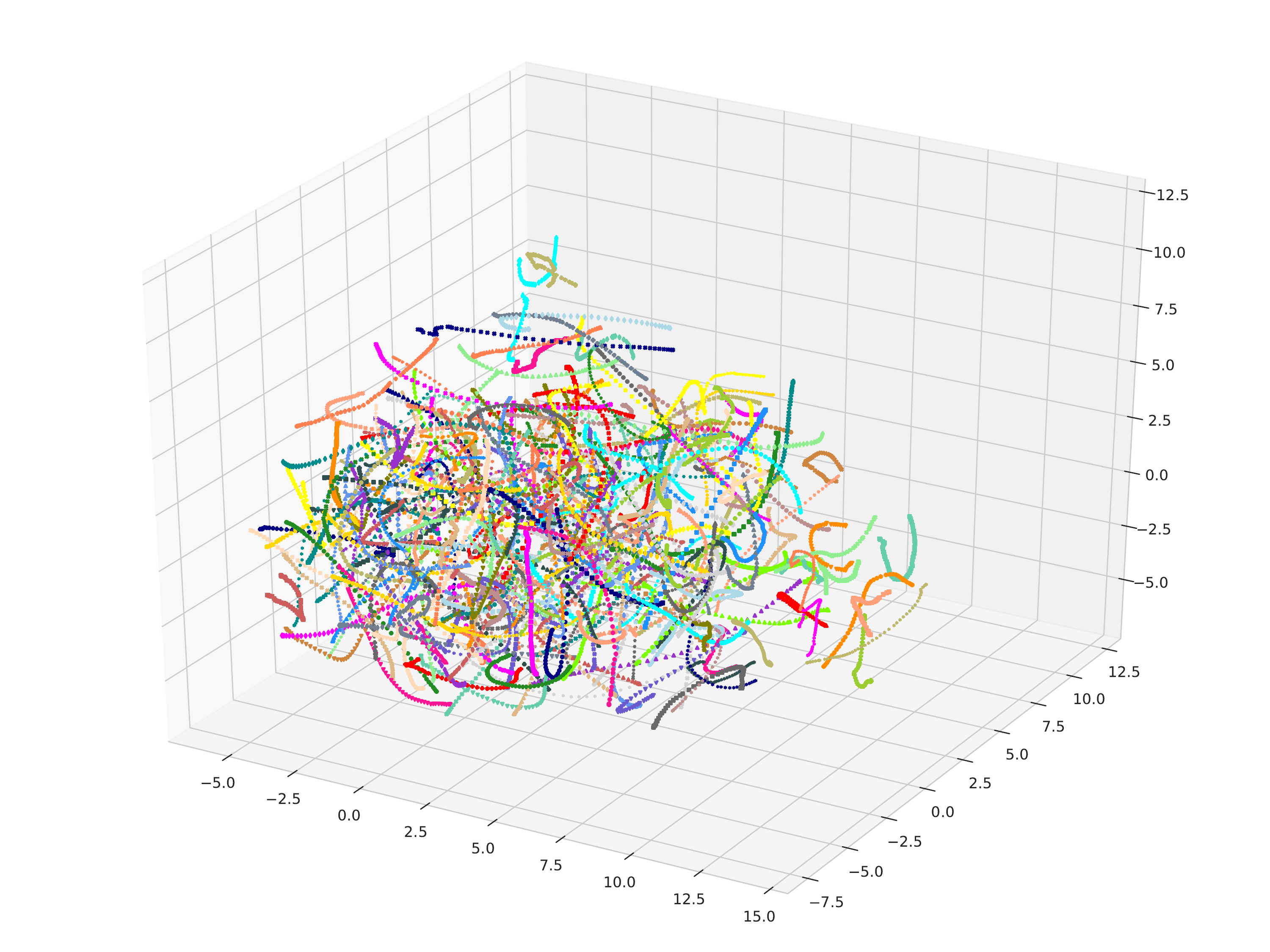}}
    \caption{UMAP of CNN weight space. Points of the same color represent the same network at different epochs. }
    \label{fig:weight_space_CNN_umap}
\end{figure}

In Figures \ref{fig:weight_space_MLP_umap} and \ref{fig:weight_space_CNN_umap}, the UMAP representation of the weight space as reduction\footnote{neighbors=3, min. distance = 0.3} of the full weight space are visualized. The CNN weight space shows smooth trajectories of nets over epochs, but they appear quite clustered. In the MLP weight space, trajectories are not as smooth, but still recognizable.
%
%

To evaluate how robust the signal of trajectories in weight space is, we investigate how much of the structure can be preserved in lower-dimensional spaces, we apply a truncated Singular Value Decomposition (SVD) to map the weight space to $50$ dimensions and apply UMAP on the reduced space for visualization.
\begin{figure}[ht]
    \centering
    \centerline{\includegraphics[width=\columnwidth]{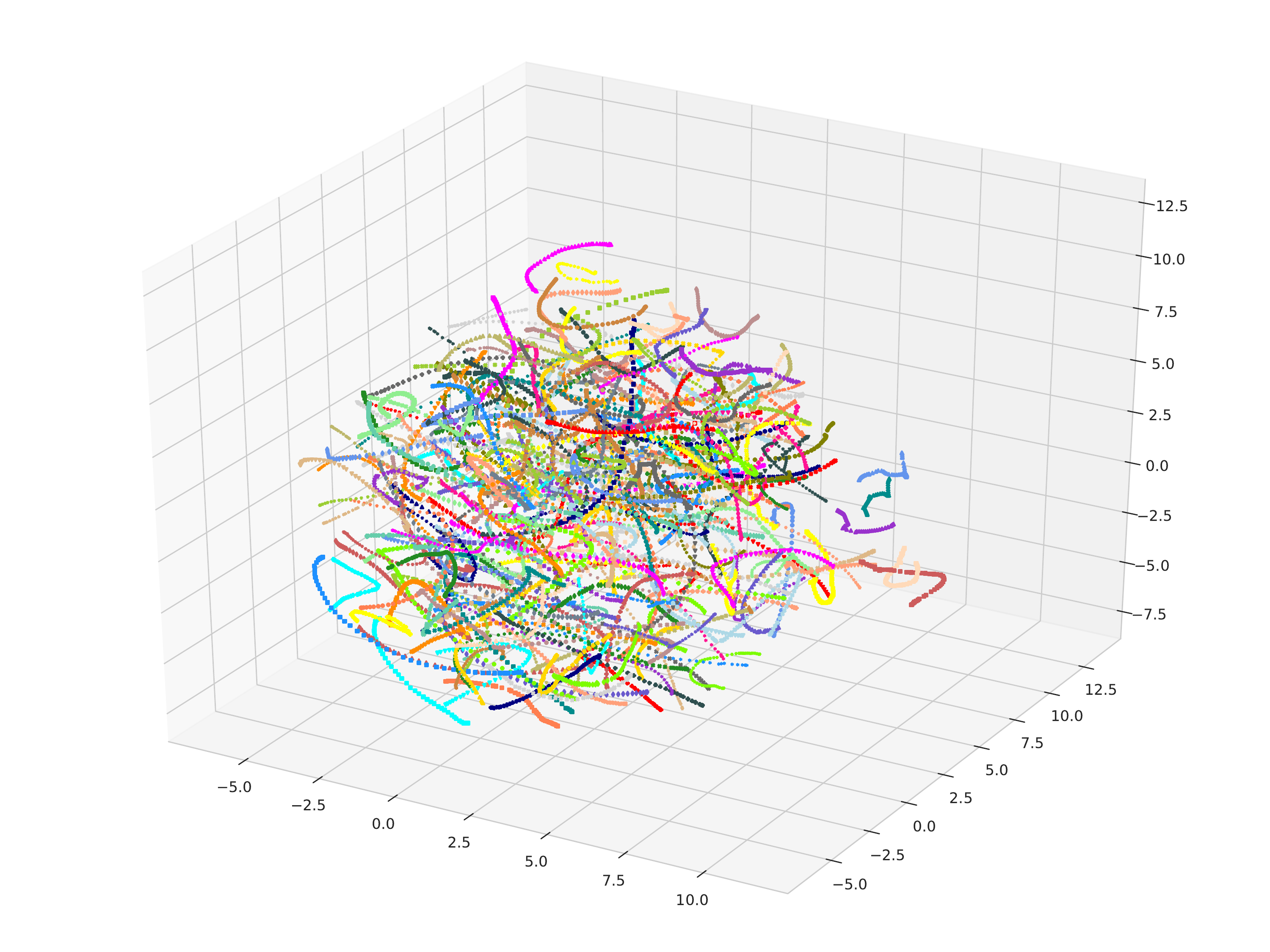}}
    \caption{UMAP of CNN SVD reduction to 50 dimensions. Points of the same color represent the same network at different epochs. }
    \label{fig:svd_umap}
\end{figure}
%
As illustrated in Figure \ref{fig:svd_umap}, CNN trajectories are cleanly separated even in the heavily compressed space. On the trajectories, larger distances are visible at initial epochs and smaller towards the end of the model training.
\begin{table*}[ht]
\caption{Quantitative Results of MLP Zoo}
\label{tab:res_MLP_zoo}
\begin{center}
\begin{tabular}{l|c|c|c}
Space / Dimension & Cluster Purity & \# of ordered initial epochs & \# of ordered epochs\\
2160 (Weight Space) & 1.0 & 4.512 & 0.357 \\
500 (SVD) & 1.0 & 4.632 & 0.339 \\
100 (SVD) & 1.0 & 4.532 & 0.325 \\
50 (SVD) & 1.0 & 4.212 & 0.303 \\
25 (SVD) & 1.0 & 3.684 & 0.273 \\
10 (SVD) & 1.0 & 2.916 & 0.227 \\
5 (SVD) & 0.843 & 2.038 & 0.177 \\
\end{tabular}
\end{center}
\end{table*}
%
%
\begin{table*}[ht]
\caption{Quantitative Results of CNN Zoo}
\label{tab:res_CNN_zoo}
\begin{center}
\begin{tabular}{l|c|c|c}
Space / Dimension & Cluster Purity & \# of ordered initial epochs & \% of ordered epochs\\
1376 (Weight Space) & 1.0 & 20.184 & 0.693 \\
500 (SVD) & 1.0 & 21.848 & 0.732 \\
100 (SVD) & 1.0 & 18.632 & 0.689 \\
50 (SVD) & 1.0 & 17.728 & 0.675 \\
25 (SVD) & 1.0 & 15.852 & 0.643 \\
10 (SVD) & 0.896 & 12.094 & 0.568 \\
5 (SVD) & 0.465 & 8.175 & 0.471 \\
\end{tabular}
\end{center}
\end{table*}
\subsection{Epoch-to-Model Assignment}
\begin{figure*}[ht]
 \centering
    \centerline{\includegraphics[width=2\columnwidth]{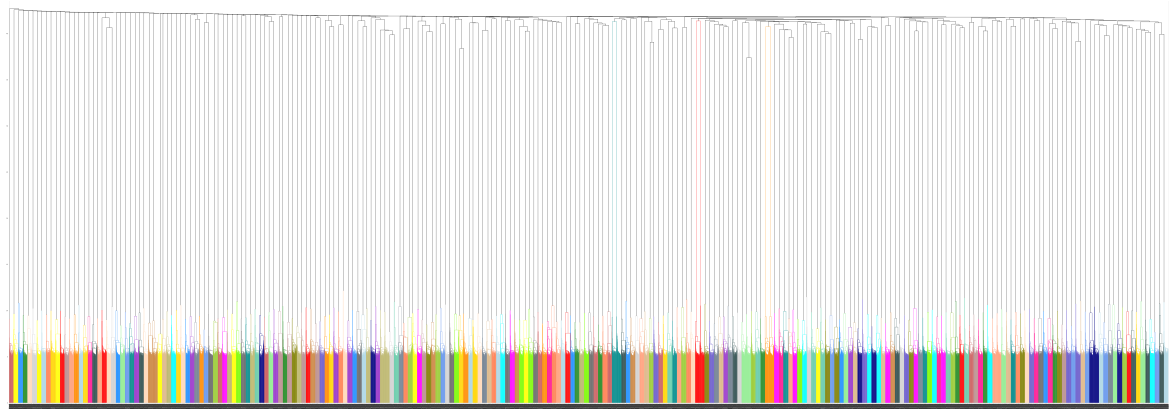}}
    \caption{Result of the agglomerative clustering of MLP model checkpoints in weight space visualized as a dendrogram. Epochs of the same DNN model, which are correctly clustered are colored in the same color. The distance between samples of a cluster is small, while the distance between model clusters is large, and almost constant over all clusters. This underlines the sparsity of the parameter space.}
\label{fig:dendrograms_full_MLP_apdx}
\end{figure*}

\begin{figure*}[ht]
 \centering
    \centerline{\includegraphics[width=2\columnwidth]{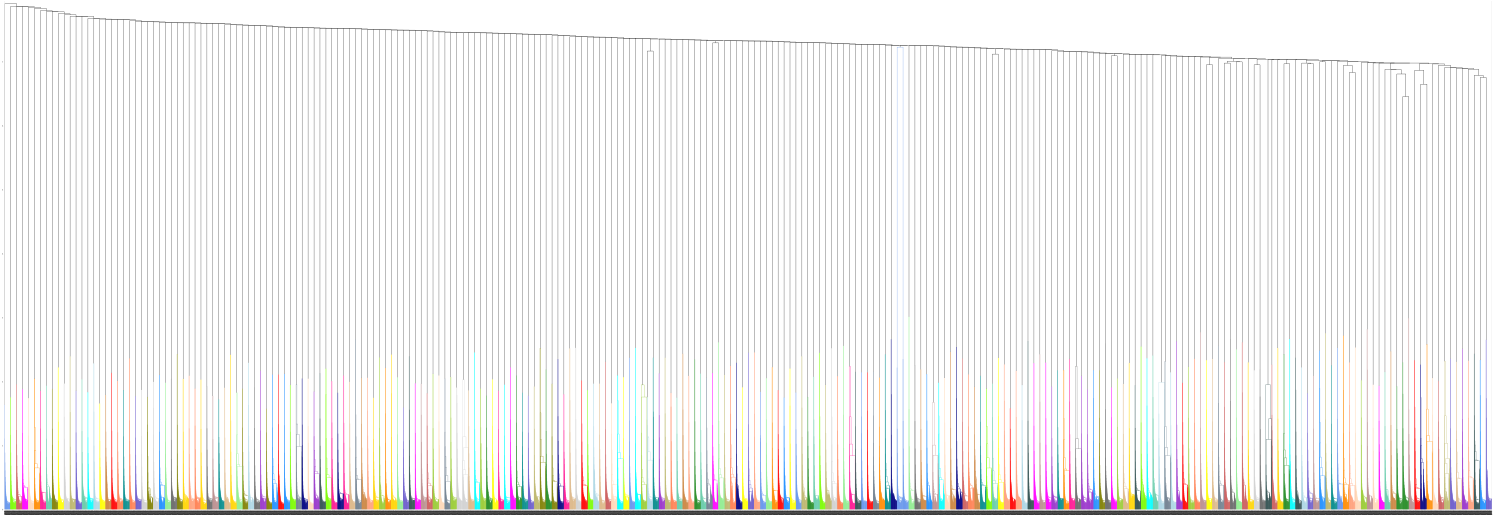}}
    \caption{Result of the agglomerative clustering of CNN model checkpoints in weight space visualized as a dendrogram. Epochs of the same DNN model, which are correctly clustered are colored in the same color. The distance between samples of a cluster is small, while the distance between model clusters is large, and very even over all clusters. This underlines the sparsity of the parameter space.}
\label{fig:dendrograms_full_CNN_apdx}
\end{figure*}

To demonstrate the representation's potential to be used as a feature version control and DNN model identification, we evaluate the weight space's capability to link single epochs (i.e. individual samples) to model trajectories.
For each architecture, all 250 models have been clustered and visualized in form of a dendrogram. Details of the agglomerative clustering (Sec.\ref{approach:structure}) are shown in Figure~\ref{fig:dendrograms}. The full dendrograms of the MLP weight space and CNN weight space can be found in Figures \ref{fig:dendrograms_full_MLP_apdx} and \ref{fig:dendrograms_full_CNN_apdx}, respectively.

We make two observations in the full dendrograms: (i) the distance between the model clusters - each of the cluster contains exactly one trajectory - is large compared to the distances within the cluster. The mutual distances between the clusters is almost constant, which we explain with the high dimensionality of the parameter space. This observation strongly underlines the sparsity of the parameter space, even for such - by today's standards - small models; (ii) the distance within the clusters is noticeably smaller than between clusters. Without considering smoothness, the euclidean distance is sufficient to identify model identities. That also reveals, that models in our experiments develop locally in weight space, i.e. they do not cross the entire weight space. While learned weight structure patterns between the models may be similar, they develop in a distributed fashion close to the initialization point. Some of that observation may be explained with the distinct non-identifiability of a solution to the training problem of DNNs.

On top of the separability, the CNN weight space appears to be ordered in the resulting dendrogram structure. The saw-tooth pattern of mergers within one cluster fits usual training patterns: large loss, gradients and weight updates in initial epochs, smaller loss and corresponding updates in later epochs.
On reduced spaces, the distance from one trajectory to another is sometimes just marginally larger than from the first epoch to the second. Here, single-linkage clustering appears to be too coarse to cover the trend and smoothness of training trajectories.

The MLP dendrogram confirms the visualization. While the individual DNN models can be separated, their epoch training order cannot always be reconstructed from the distances in weight space as seen in, see detail in Figure ~\ref{fig:dendrograms}. The distance between the last epochs is not small, indicating that either a) the nets have not converged or b) the weight space contains noise-like change in weights driven by non-zero loss that doesn't change nets behavior at the end of the training.

Tables ~\ref{tab:res_CNN_zoo} and ~\ref{tab:res_MLP_zoo} show the results of the quantitative evaluation, which confirm the dendrograms. For the CNN zoo, clusters remain pure and uphold good order even if the dimensionality is reduced to 50. Below, both purity as well as ordering decline. The MLP zoo can be clustered to high purity even on SVD 10, but the ordering is very limited. In some limits, the mutual distance between model checkpoints is sufficient to recover a model's training history.

We would like to emphasize that our model zoo only differs in their seeds, so the models and their trajectories are expected to be similar. Nevertheless, the weight space shows to have structure rich enough to be used as feature to track the training development of models.

%
%
%
\section{Conclusion}
\label{conclusion}
Our experiments show promising results to facilitate DNN monitoring directly on the weight space. The weight space of zoos of small models show training trajectories which a) are smooth, b) relate training update to distance and c) facilitate epoch-to-model assignment and d) allow to tell different models apart. \\
We suspect that curvature and step-size over epochs along the trajectories are shaped by the data the models are trained on over the mini-batches, and so may reveal domain shifts in the training data. These may be used to detect adversarial attacks, or a normal change in observations. As the model behavior becomes more volatile and unpredictable when exposed to a different domain, these situations are crucial to single out in ML monitoring. Future work will focus on investigating the trajectories and their properties further.\\
Lastly, the combinatorial small chance of coinciding allows for model trajectories to be interpreted as model IDs, and as such relevant for intellectual property protection. Furthermore, if models are fine-tuned or transfer learned to new tasks, they would show up as non-smooth branches forking off of the main trajectory, facilitating model provenance.

\bibliography{bibliography}

\end{document}